\documentclass{article}

\usepackage{amsmath}
\usepackage{graphicx}
\usepackage[colorlinks=true, allcolors=blue]{hyperref}
\usepackage{xcolor}
\usepackage{ amssymb }
\usepackage{amsthm}
\usepackage{multirow}
\usepackage{makecell}
\usepackage{enumitem}

\usepackage{url}
\usepackage{titlesec}
\usepackage{subcaption}
\usepackage{booktabs}
\usepackage{siunitx}
\titleformat{\paragraph}
{\normalfont\normalsize\bfseries}{\theparagraph}{1em}{}
\titlespacing*{\paragraph}
{0pt}{3.25ex plus 1ex minus .2ex}{1.5ex plus .2ex}

\usepackage{pgfplots}
\pgfplotsset{compat=1.18}
\usepackage{lastpage}
\usepackage{fancyhdr}

\title{Delete and Retain: Efficient Unlearning for Document Classification}
\author{
Aadya Goel$^{1}$ \qquad
Mayuri Sridhar$^{2}$\\
\\[-1em]
{\small $^{1}$Acton-Boxborough Regional High School} 
{\small $^{2}$MIT}
}
\date {\small   }

\begin{document}
\maketitle

\begin{abstract}
\noindent Machine unlearning aims to efficiently remove the influence of specific training data from a model without full retraining.  While much progress has been made in unlearning for LLMs, document classification models remain relatively understudied. In this paper, we study class-level unlearning for document classifiers and present Hessian Reassignment, a two-step, model-agnostic solution. First, we perform a single influence-style update that subtracts the contribution of all training points from the target class by solving a Hessian–vector system with conjugate gradients, requiring only gradient and Hessian-vector products. Second, in contrast to common unlearning baselines that randomly reclassify deleted-class samples, we enforce a decision-space guarantee via Top-1 classification. On standard text benchmarks, Hessian Reassignment achieves retained-class accuracy close to full retrain-without-class while running orders of magnitude faster. Additionally, it consistently lowers membership-inference advantage on the removed class, measured with pooled multi-shadow attacks. These results demonstrate a practical, principled path to efficient class unlearning in document classification. \vspace{1em}

\noindent \textbf{Keywords:} Machine Unlearning · Document Classification · Hessian · Data Privacy · Forgetting 
\end{abstract}

\section{Introduction}
Modern machine learning systems increasingly operate under legal and user-driven deletion requests, including \textit{right-to-be-forgotten} \cite{SW19} provisions in regulations such as CCPA \cite{BT2} and GDPR \cite{GDPR}. Exact unlearning by retraining from scratch is conceptually simple but often impractical for production models due to compute budgets and continual data churn. This tension has motivated \emph{machine unlearning} \cite{CY15}, algorithms that remove the influence of specified training data from a deployed model without full retraining \cite{BL21,GA19,NS21}. While prior work has largely considered removing individual points or small subsets via influence approximations \cite{KL17}, gradient-based adjustments \cite{NS21}, or partitioned training \cite{BL21}, these designs do not directly exploit additional \emph{structure} in the deletion request.

We study the setting where an entire class must be forgotten in a trained document classifier. This situation arises when a label is deemed sensitive, a cohort associated with a category requests erasure, or other related events. After unlearning, the model must still produce predictions over the remaining labels for all inputs, including documents that originally belonged to the deleted class.

While existing methods remove the influence of the target class, they ignore the score ordering learned by the original model, degrading model utility~\cite{KS24}. Instead, we propose unlearning through top-1 reclassification.

Our key observation is that a trained document classifier already encodes a coherent ranking over these non-target labels for each input. We design a two-step, model-agnostic procedure that preserves this structure while removing the influence of the target class. We apply a single second-order\ ``downweight''\ correction step and then do deterministic reassignment.
\vspace{1em}

\noindent This paper makes the following contributions:
\begin{itemize}
    \item We formalize \emph{class-level unlearning} for document classifiers and identify the decision-space ambiguity it introduces.
    \item It introduces \textbf{Hessian Reassignment}, a simple two-step unlearning procedure that combines a single class-level Hessian downweight update with a deterministic next-top-1 decision rule over non-target labels. 
    \item It shows empirically, on standard text benchmarks, that this procedure matches complete retraining on retained-class accuracy while substantially improving privacy and utility over Random Relabeling.. 
\end{itemize}
\section{Related Work}
\subsection{Machine Unlearning Frameworks}
Machine unlearning methods aim to remove the influence of specific training data without retraining from scratch. Several lines of work study scalable exact procedures for deep models \cite{SC25}, feature- and label-level removal \cite{AW23}, and the utility–complexity trade-offs of unlearning under distribution shift \cite{YA25}. Other work emphasizes practical challenges around data quality, storage, and access, treating unlearning as a systems and data-management problem as much as an algorithmic one \cite{MX24}. 

In language models, unlearning has been used for privacy-preserving fine-tuning \cite{MR25}, probing what information remains after deletion \cite{AD25}, analyzing the impact of pretraining data via removal \cite{YZ24}, and efficiently forgetting specific content or behaviors in large pretrained models \cite{JC23,JY24}. Overall, these frameworks primarily address point-level or subset-level deletion in deep architectures. In contrast, this paper considers removing an entire class from a document classifier while preserving the structure of predictions over the remaining labels.

\subsection{Training Data Influence and Second-Order Methods}

Training-data influence methods use information about the local \emph{curvature} of the loss landscape, captured by the Hessian, to approximate how adding or removing examples would shift model parameters. One line of work estimates influence by tracing gradient descent trajectories \cite{GP20}, while another develops second-order group influence functions that analyze how sets of examples affect predictions via Hessian-based approximations \cite{SB20}. Subsequent studies show that such estimates can be fragile in deep models and survey techniques for more robust influence analysis across architectures \cite{SB21,ZH24}. 

Influence-style ideas have also been applied directly to unlearning, for example by tracing influential training data in language models \cite{MI24} or using differential gradient approximations to scale influence estimation to larger systems \cite{HT25}. The approach in this paper is conceptually related in that it uses second-order information, but rather than estimating pointwise influence or tracing training trajectories, it applies a single class-level Hessian downweight update that removes the aggregate contribution of all deleted-class examples in one step.

\subsection{Unlearning in NLP and Document Models}
Beyond generic unlearning frameworks, there is a growing body of work focused on natural language processing and large language models. Several methods use machine unlearning to remove memorized content, sensitive information, or undesired behaviors from pretrained language models \cite{MR25, YZ24, JC23, JY24}. These approaches often operate at the level of model parameters, adapters, or task-specific heads for sequence models, and are typically designed for transformer-based architectures.

The setting studied in this paper is different: classical document classification with TF--IDF features and a linear-softmax head. While the techniques developed for language models could in principle be adapted to simpler pipelines, existing methods do not exploit the specific structure of class-level deletions in multi-class document classifiers. Here, the goal is to remove a full label and all of its training examples while preserving the model’s learned ordering over the remaining labels. Unlike point deletions, class-level unlearning simultaneously alters the empirical label distribution and the geometry of multiple decision boundaries. 

A central design choice here is how the model should behave on deleted-class documents. A simple and commonly referenced approach in document classification is to \emph{randomly reassign} removed samples across the remaining classes (``Random Relabeling'') \cite{KS24}. Although easy to implement, this destroys the score ordering learned by the original model, often degrading utility and inducing unstable decision regions. Our proposed method addresses this by combining a deterministic next-top-1 decision rule with a single Hessian-based update.

\subsection{Membership Inference Attack}
Membership inference attacks (MIAs) evaluate whether a model reveals if a particular data point was part of its training dataset. A central framework is the \emph{shadow-model} approach, in which an adversary trains auxiliary models to mimic a held-out target model and then uses confidence scores to distinguish members from non-members \cite{RS17}. Follow-up work generalizes these ideas to a broader range of models and data regimes and shows that overfitting strongly correlates with privacy risk, and that membership inference remains a concern even for synthetic or heavily processed data \cite{AS19,SY18,ZZ21}. 

These results motivate our privacy metrics. By measuring membership-inference ROC--AUC on both retained-class and deleted-class examples, we can quantify how much the proposed unlearning procedure reduces membership leakage, especially for the class that is supposed to be “forgotten.”

\section{Problem Formulation}
\textbf{Setting and notation.} 
Let $\mathcal{D}=\{(x_i,y_i)\}_{i=1}^n$ be a training corpus with documents $x_i\in\mathcal{X}$ and labels $y_i\in\mathcal{Y}=\{1,\dots,K\}$.
A classifier with parameters $W$ produces class probabilities
\[
p_W(y\mid x)=\mathrm{softmax}\!\big(s_W(x)\big)_y,
\]
where $s_W(x)\in\mathbb{R}^K$ are model scores (architecture-agnostic). The empirical objective is
\[
\ell(W;\mathcal{D})=\frac{1}{n}\sum_{i=1}^n \mathcal{L}\!\big(s_W(x_i),y_i\big)+\Omega(W),
\]
with a standard loss $\mathcal{L}$ (e.g., cross-entropy) and regularizer $\Omega$. Let the following denote the trained model $$W^\star=\arg\min_W \ell(W;\mathcal{D}).$$
\textbf{Class-level deletion.}
Given a target class $c\in\mathcal{Y}$, the request is to \emph{forget} all its training examples
$\mathcal{D}_c=\{(x_i,y_i):y_i=c\}$.
\vspace{1em}

\noindent The retained dataset and label set are
$\mathcal{D}_{-c}=\mathcal{D}\setminus\mathcal{D}_c$ and
$\mathcal{Y}_{-c}=\mathcal{Y}\setminus\{c\}$, respectively.
\vspace{1em}

\noindent \textbf{Unlearning objective.}
An unlearning operator $\mathcal{U}_c$ maps $(\mathcal{D},W^\star)$ to unlearned model $\widetilde{W}$ such that:
\begin{enumerate}
    \item \textbf{Deletion consistency.} On any evaluation set supported on $\mathcal{Y}_{-c}$,
    the performance of $\widetilde{W}$ matches that of the \emph{complete retraining} solution
    \[
    W_{-c}^\star=\arg\min_W \ell(W;\mathcal{D}_{-c}).
    \]
    \item \textbf{Viable predictions for removed examples.}
    For all $x$ originally labeled $c$, $\widetilde{W}$ produces a distribution over $\mathcal{Y}_{-c}$ (i.e., does not route probability mass to the removed class).
    \item \textbf{Privacy after deletion.}
    Membership information about $\mathcal{D}_c$ is obscured under standard attacks (e.g., MIA), ideally approaching the attack performance against $W_{-c}^\star$.
\end{enumerate}
We measure (1) via accuracy on $\mathcal{Y}_{-c}$, (2) via probability diagnostics over $\mathcal{Y}_{-c}$ (margins, KL), and (3) via removed-class MIA AUC.
\vspace{1em}

\noindent \textbf{Structure-Preserving Reassignment via Next Top-1}
For each deleted-class document $x\in\mathcal{D}_c$, define the \emph{next top-1} label under the \emph{pre-unlearning} model $W^\star$:
\[
\phi_{-c}(x) \;=\; \arg\max_{y\in\mathcal{Y}_{-c}} \, p_{W^\star}(y\mid x).
\]
This answers the counterfactual: \emph{If class $c$ had never existed, which class would the deleted-class document be classified to?} We refer to $x\mapsto\phi_{-c}(x)$ as a \emph{structure-preserving reassignment}, since it preserves the learned score ordering among non-removed classes.
\vspace{1em}

\noindent \textbf{Golden Standard}
We treat $W_{-c}^\star$ as the ``Golden Standard'' reference. The ideal unlearned model $W_{-c}$ is defined as the classifier retrained from scratch on $D_{-c}$. 
\vspace{1em}

\noindent For a test distribution $\mathcal{T}$ over $\mathcal{X}\times\mathcal{Y}_{-c}$, a utility metric $U$ (e.g., accuracy) and a privacy metric $P$ (e.g., MIA AUC on $\mathcal{D}_c$), we desire
\[
U(\widetilde{W};\mathcal{T}) \approx U(W_{-c}^\star;\mathcal{T})
\quad\text{and}\quad
P(\widetilde{W}) \approx P(W_{-c}^\star).
\]
\noindent \textbf{State-of-the-Art Baseline: Random Relabeling}
The current state-of-the-art class-level baseline in document classification unlearning is \emph{random relabeling}~\cite{KS24}, where each deleted-class
document $x \in \mathcal{D}_c$ is reassigned a label sampled uniformly from the remaining set $\mathcal{Y}_{-c}$ and the model is fine-tuned or retrained on the modified dataset.

By construction, this scheme treats all non-$c$ labels as equally plausible for deleted examples, ignoring the model’s pre-existing ranking over $\mathcal{Y}_{-c}$ and replacing it with independent random targets. In other words, it injects heavy label noise concentrated on the forget set.

Prior work has noted that such relabeling-based unlearning can impair generalization or cause over-forgetting in deep models~\cite{KS24,NM25, LG23}, and in our experiments (Section~\ref{sec:experiments}) we likewise observe that random relabeling often trades away retained-class utility to achieve privacy on the
deleted class.
\vspace{1em}

\noindent \textbf{Requirements Specific to Class-Level Deletions}
\begin{enumerate}
    \item \textbf{Order preservation.} For $x\notin\mathcal{D}_c$, the ranking over $\mathcal{Y}_{-c}$ under $\widetilde{W}$ remains close to that under $W^\star$.
    \item \textbf{Privacy focus.} Post-deletion attack advantage on $\mathcal{D}_c$ should approach chance (AUC~$\approx0.5$), while retaining utility on $\mathcal{D}_{-c}$.
\end{enumerate}
These properties define the target behavior independently of architecture and guide both our method and evaluation.

\section{Methodology}
\label{sec:method}

Let $\mathcal{D}=\{(x_i,y_i)\}_{i=1}^n$ be a labeled corpus with $y_i\in\mathcal{Y}$ and model parameters $\theta$ trained to minimize a regularized empirical loss $$\mathcal{L}(\theta)=\sum_{i=1}^n \ell(x_i,y_i;\theta)+\Omega(\theta).$$ A deletion request specifies a target class $c\in\mathcal{Y}$. After unlearning, the model should (i) forget membership information tied to class $c$ and (ii) preserve predictions among the remaining labels $\mathcal{Y}_{-c}$. We report utility as test accuracy excluding class $c$, and privacy via retained-class membership-inference AUC.

\subsection{Top-1 classification reassignment}
For each training point with $y_i=c$, define its next top-1 label under the pre-unlearning model:
\[
r_i \;=\; \arg\max_{k\neq c} s_k(x_i;\theta),
\]
where $s_k$ are the model scores (e.g., logits). This deterministic rule preserves the model’s score ordering among non-$c$ classes and provides a stable target after deletion. We use this rule to define target reassignment behavior for deleted-class points.

\subsection{Hessian downweight update}
Recall the regularized empirical loss $\nabla \mathcal{L}(\theta)$. Deleting class $c$ replaces $\mathcal{L}$ by
\[
\mathcal{L}^{(-c)}(\theta)
= \sum_{i\notin \mathcal{I}_c}\!\ell(x_i,y_i;\theta) + \Omega(\theta),
\qquad
\mathcal{I}_c 
:=\{\,i:\, y_i=c\,\}.
\]
The new optimizer $\theta^\star$ satisfies $$\nabla \mathcal{L}^{(-c)}(\theta^\star)=0.$$
A second-order Taylor expansion of $\nabla \mathcal{L}^{(-c)}$ around $\theta$ gives
\[
\nabla \mathcal{L}^{(-c)}(\theta^\star)
\approx \nabla \mathcal{L}^{(-c)}(\theta) + H(\theta)\,(\theta^\star-\theta)
= -\,g_c(\theta) + H(\theta)\,\Delta \theta,
\]

\noindent where $H(\theta) := \nabla^2 \mathcal{L}(\theta)$\footnote{Using $H(\theta)$ rather than $\nabla^2\mathcal{L}^{(-c)}(\theta)$ yields the same first-order correction at $\theta$ and is more stable in practice.} and
\[
g_c(\theta) := \sum_{i\in \mathcal{I}_c}\nabla \ell(x_i,y_i;\theta)
\]
is exactly the gradient contribution of the deleted class. Setting the linearized gradient to zero yields the \emph{inexact Newton} step
\begin{equation}
\label{eq:newton}
\Delta \theta \;=\; H(\theta)^{-1} g_c(\theta),
\qquad
\theta' \;=\; \theta \;-\; \Delta \theta.
\end{equation}
We call Eq.~(\ref{eq:newton}) a \emph{Hessian downweight}. It subtracts (downweights) the first- and second-order influence of the deleted terms from the optimality conditions without performing complete retraining.

In Newton methods, $H(\theta)^{-1}$ maps a desired change in the gradient (\emph{here}, removing $g_c$) to a parameter move that cancels it to first order. Intuitively, $H^{-1}$ rescales each parameter direction by the local curvature: flat directions (small eigenvalues) require larger moves; stiff directions require smaller ones. Under a second-order optimization, $\theta'$ is the first order approximation to the optimizer obtained by retraining on $\mathcal{D}_{-c}$.

\subsection{Instantiation for multinomial logistic regression}
For a linear-softmax classifier with weights $W\in\mathbb{R}^{K\times d}$ (no intercept), let $P=\mathrm{softmax}(XW^\top)$. The gradient contribution of one sample $(x_i,y_i)$ is
\[
\nabla_W \ell_i \;=\; (p_i - e_{y_i})\, x_i^\top,
\]
and the Hessian-vector product with $V\in\mathbb{R}^{K\times d}$ is
\[
H[V]
\;=\;
\Big( \big(P \odot U\big) - \big(P \odot (s\,\mathbf{1}_K^\top)\big) \Big)^{\!\top} X
\;+\; \lambda V,
\]
where $$U = XV^\top \in \mathbb{R}^{n\times K}$$ and $s\in\mathbb{R}^n$ has entries
$$s_i = \sum_{k=1}^{K} P_{ik}\,U_{ik}$$
(i.e., the per-row sum over classes). Here $\lambda$ comes from the $\ell_2$ regularizer. We compute $\Delta = H^{-1}G_c$ via conjugate gradients using only HVPs, and set $W' = W - \Delta$.

\subsection{Zeroing the removed output at release}
At deployment we make the removed class inert by nulling its output channel (e.g., zeroing the corresponding logit row in a linear-softmax head) and re-normalizing probabilities over $\mathcal{Y}_{-c}$. This prevents any downstream consumer from selecting class $c$.

\section{Baselines \& Evaluations}
\label{sec:baseline}

\subsection{Baselines}
\textbf{Hessian Reassignment} We will refer to our algorithm as Hessian Reassignment throughout the rest of this paper. This represents the Hessian downweight update (Sec.~\ref{sec:method}) with top-1 classification reassignment and a zeroed output for the removed class at release.
\vspace{1em}

\noindent \textbf{Random Relabeling (state-of-the-art baseline).}
All training points with $y_i=c$ are reassigned uniformly at random to labels in $\mathcal{Y}_{-c}$, after which the model is fine-tuned/updated on the modified corpus. This baseline ignores the original score ordering and serves as the method we aim to outperform. Note that random relabeling keeps all datapoints with $y_i=c$ and serves as an approximate unlearning mechanism rather than a deletion-based ground truth.
\vspace{1em}

\noindent \textbf{Complete retraining (Golden Standard).} 
We remove all training examples with label $c$ and retrain the classifier 
from scratch on $D_{-c}$. This produces the ideal counterfactual model $W_{-c}$ which represents the exact behavior the system would exhibit had class $c$ never been part of the training set.
\subsection{Metrics}
\subsubsection{Model Utility}
\label{sec:metric}
We report two accuracy-style utilities:
\vspace{1em}

\noindent \textbf{Accuracy excluding the removed class.}
Let $\widehat{y}(x)$ be the post-unlearning prediction over labels $\mathcal{Y}_{-c}$ (the removed output is made inert and probabilities re-normalized). We compute
\[
\mathrm{Acc}_{\setminus c}
=\frac{1}{|\{i:\, y_i\neq c\}|}\sum_{i:\,y_i\neq c}\mathbf{1}\{\widehat{y}(x_i)=y_i\}.
\]

\noindent \textbf{Agreement on the removed class w.r.t.\ the golden standard.}
Let $\mathcal{I}_c=\{i:\,y_i=c\}$ be the indices of the deleted class. Let $\widehat{y}^{\,\text{ours}}(x)$ denote our model’s prediction after unlearning and $\widehat{y}^{\,\text{GS}}(x)$ denote the prediction from \emph{complete retraining} model $W_{-c}$ (Golden Standard). We measure how well our method reproduces
the golden standard on the removed-class documents:
\[
\mathrm{Agree}_c
=\frac{1}{|\mathcal{I}_c|}\sum_{i\in\mathcal{I}_c}
\mathbf{1}\!\left\{\widehat{y}^{\,\text{ours}}(x_i)=\widehat{y}^{\,\text{GS}}(x_i)\right\}.
\]

\subsubsection{Privacy (Membership Inference).}
\label{sec: mia}
We evaluate membership–inference risk with a fixed attacker trained on shadow models. For a target model after removing class $c$, let $p(x)\in\Delta^{K-1}$ denote the prediction vector over the remaining classes (i.e., we drop class $c$ and re-normalize). The attacker uses a feature vector $\phi(x)$ that concatenates several statistics derived from the model’s prediction on input $x$. Let $\Vert$ denote vector concatenation. Then
\[
\phi(x)
\;=\;
\bigl[\, p(x) \;\Vert\; H\!\bigl(p(x)\bigr) \;\Vert\; -\log p_{\tilde y}(x) \;\Vert\; \mathrm{gap}\!\bigl(p(x)\bigr) \,\bigr],
\]
where entropy is $$H(p)=-\sum_j p_j\log p_j,$$ $\tilde y$ is the model’s top-1 predicted label among the retained classes, and the top-2 probability gap is $$\mathrm{gap}(p)=p_{(1)}-p_{(2)}$$ 
($p_{(1)}\ge p_{(2)}$ are the largest two entries of $p$). 
\vspace{1em}

\noindent We instantiate a membership attacker as a scoring function
\[
f(x;\phi) \in [0,1],
\]
which takes the feature vector $\phi(x)$ as input and outputs a membership score (higher values indicating that $x$ is more likely to be a training example). Concretely, $f(x;\phi)$ is implemented as a logistic regression classifier trained on shadow data to distinguish \emph{member} (train) vs.\ \emph{non\-member} (test) examples using $\phi(x)$ as features.
\vspace{1em}

\noindent We then report ROC–AUC on the target model under two filters:
\[
\text{AUC}_{\mathrm{ret}} \;=\; \mathrm{ROC\text{-}AUC}\bigl(\{i: y_i\neq c\}\bigr),
\qquad
\text{AUC}_{c} \;=\; \mathrm{ROC\text{-}AUC}\bigl(\{i: y_i=c\}\bigr).
\]
Lower ROC-AUC scores indicate a stronger privacy guarantee. In particular we desire $\text{AUC}_{c}\approx 0.5$.

\subsection{Datasets and Experimental Setup}
\label{sec:datasets-setup}
\textbf{Datasets.}
We evaluate on three widely used multi-class document classification benchmarks: \textsc{20 Newsgroups}, \textsc{AG News}, and \textsc{DBPedia-14}. These datasets are standard in text classification research and offer complementary levels of difficulty and class granularity, making them well-suited for studying class-level unlearning.

Specifically, \textsc{20 Newsgroups} contains documents from 20 balanced topics spanning technology, politics, recreation, and science. Its relatively high number of classes and moderate document length make it a challenging test of whether unlearning preserves fine-grained decision boundaries.

\textsc{AG News} includes news articles labeled into 4 broad categories. Its lower intrinsic dimensionality allows us to test whether our method remains stable when classes are coarse-grained and semantically distinct, a setting where random relabeling tends to be particularly harmful.

\textsc{DBPedia} is a 14-class ontology-based dataset derived from Wikipedia, with short, well-structured descriptions. It provides a clean, high-accuracy baseline to assess whether unlearning introduces noise.

Together, these datasets cover a spectrum of class counts, label semantics, document lengths, and difficulty levels. They form a natural testbed for class-level unlearning as one class changes the structure of the remaining label space in different ways across each dataset.
\vspace{1em}

\begin{figure}[ht!]
\begin{center}
\includegraphics[scale=0.4]{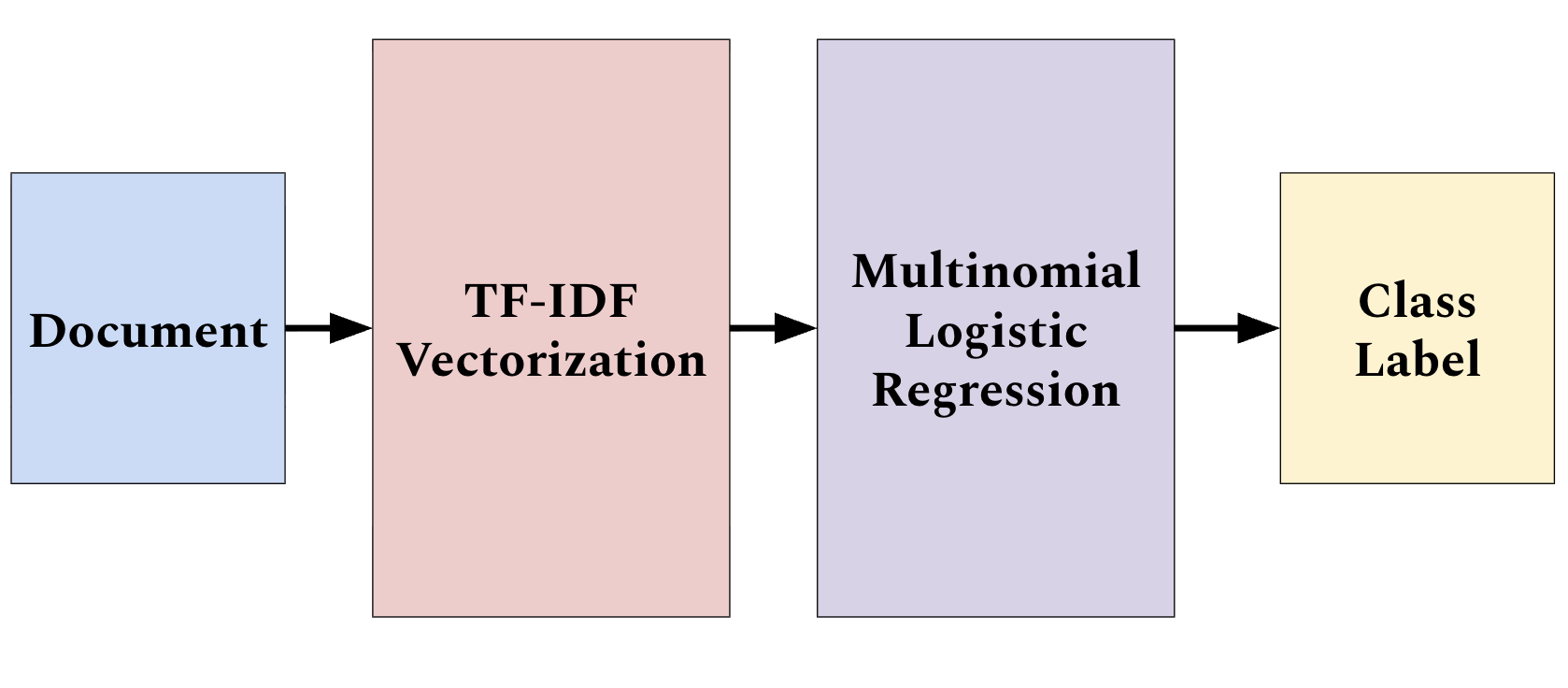}
\end{center}
\caption{Architecture of the backbone document classifier.}
\label{fig:architecture}
\end{figure}

\noindent \textbf{Model.} Our backbone classifier is a multinomial logistic regression model with $\ell_2$ regularization, trained on TF--IDF features (Fig~\ref{fig:architecture}). For each dataset, we tokenize, lowercase, remove stopwords, and construct TF--IDF representations using sublinear term frequency, \texttt{min\_df=2}, and a maximum vocabulary size of 50K. The classifier is trained with L-BFGS until convergence with a gradient
tolerance of $10^{-5}$ and regularization strength $C=10.0$ unless otherwise noted. A validation split comprising $10\%$ of the training data is used for early stopping. This simple linear-softmax pipeline achieves strong baseline accuracy while exposing the influence of curvature (Section~\ref{sec:method}) in a transparent manner.

\noindent \textbf{Unlearning Procedure.}
For each class $c$, we compute the class-level gradient term for $\mathcal{D}_c$,
solve the Hessian–vector system with conjugate gradients (tolerance $10^{-4}$,
maximum 200 iterations), apply the downweight update, and zero the deleted
channel at inference. All experiments are repeated over 5 random seeds.
\vspace{1em}

\noindent \textbf{Attack Model.}
We adopt the shadow-model membership-inference setup used in prior work. We construct a pooled shadow dataset by repeatedly sub-sampling the original training corpus into disjoint \emph{shadow\_train}/\emph{shadow\_holdout} splits and training $S=10$ shadow models, each replicating the full training pipeline. For post-unlearning evaluation, each shadow applies the same unlearning procedure (remove class $c$, apply the Hessian downweight, and zero the removed logit) before emitting prediction vectors.

For each example, we compute the attacker feature vector $\phi(x)$ defined in Section~\ref{sec: mia}, and train a logistic-regression attacker on the pooled shadow features with balanced classes. The attacker is then evaluated on the target model to obtain $\text{AUC}_{\mathrm{ret}}$ and $\text{AUC}_c$.

\section{Results} 
\label{sec:experiments}

\subsection{Model Utility}
\label{sec:utility}

We measure utility on the following conditions in Table~\ref{tab:utility}: 
\begin{enumerate}[label=(\alph*)]
    \item Overall accuracy on the pre-unlearning model
    \item Post-unlearning accuracy on the \emph{retain classes} with \textbf{Golden Standard}
    \item  Post-unlearning accuracy on the \emph{retain classes} with \textbf{Random Relabeling}
    \item Post-unlearning accuracy on the \emph{retain classes} with \textbf{Hessian Reassignment}
    \item The agreement rate on test items with $y{=}c$ between the \textbf{Golden Standard} and \textbf{Hessian Reassignment}, i.e., the fraction of deleted-class test documents for which both methods predict the same label among the retained classes.
\end{enumerate}

\begin{table}[ht!]
\centering
\caption{Utility comparison and agreement on deleted class.}
\label{tab:utility}
\begin{tabular}{
  l
  S[table-format=2.2]  % (a) Pre (overall)
  S[table-format=2.2]  % (b) Random relabel (retained)
  S[table-format=2.2]  % (c) Golden std. (retained)
  S[table-format=2.2]  % (d) Our algorithm (retained)
  S[table-format=2.2]  % (e) Agreement on deleted class
}
\toprule
\multicolumn{1}{c}{\textbf{Method}} &
\multicolumn{1}{c}{20 Newsgroup} &
\multicolumn{1}{c}{AG News} &
\multicolumn{1}{c}{DBPedia-14} \\
\midrule
Pre-Unlearning & 93.58\% & 92.14\% & 98.33\% \\
Golden Standard & 93.73\% & 95.65\% & 98.43\% \\ 
Random Relabel & 83.45\% & 84.58\% & 88.46\% \\ 
Hessian Reassignment & 93.15\% & 95.04\% & 97.55\% \\ 
Unlearned Class Agreement & 87.77\% & 88.16\% & 90.46\% \\ 
\bottomrule
\end{tabular}
\end{table}

\noindent Across all three datasets, \textbf{Hessian Reassignment} matches the \textbf{Golden Standard} on retained-class accuracy while substantially outperforming \textbf{Random Relabeling}. The agreement on the deleted class is high ($\approx$88\%), indicating that the top-1 reassignment used in \textbf{Hessian Reassignment} is closely mirrored by the \textbf{Golden Standard}. 

\subsection{Confidence Distribution}
To understand how unlearning affects the model’s behavior, we study changes in \emph{confidence}. For a classifier with output probabilities $p(x)$, the confidence that the model assigns to a label $y$ is captured by the \emph{top-1 probability margin}
\[
m(x; y) \;=\; p_{y}(x)\;-\;\max_{j\neq y} p_j(x).
\]
\noindent A positive margin means the model prefers $y$ over all alternatives (\textbf{confident correct prediction}); a negative margin means some other label has higher probability (\textbf{confident misprediction}). Values near $1$ correspond to near-total confidence, values near $0$ to uncertainty, and values near $-1$ to high confidence in a wrong label.
\vspace{1em}

\begin{figure}[hbt!]
\begin{subfigure}[t]{.5\textwidth}
    \centering
    \includegraphics[width=0.95\linewidth]{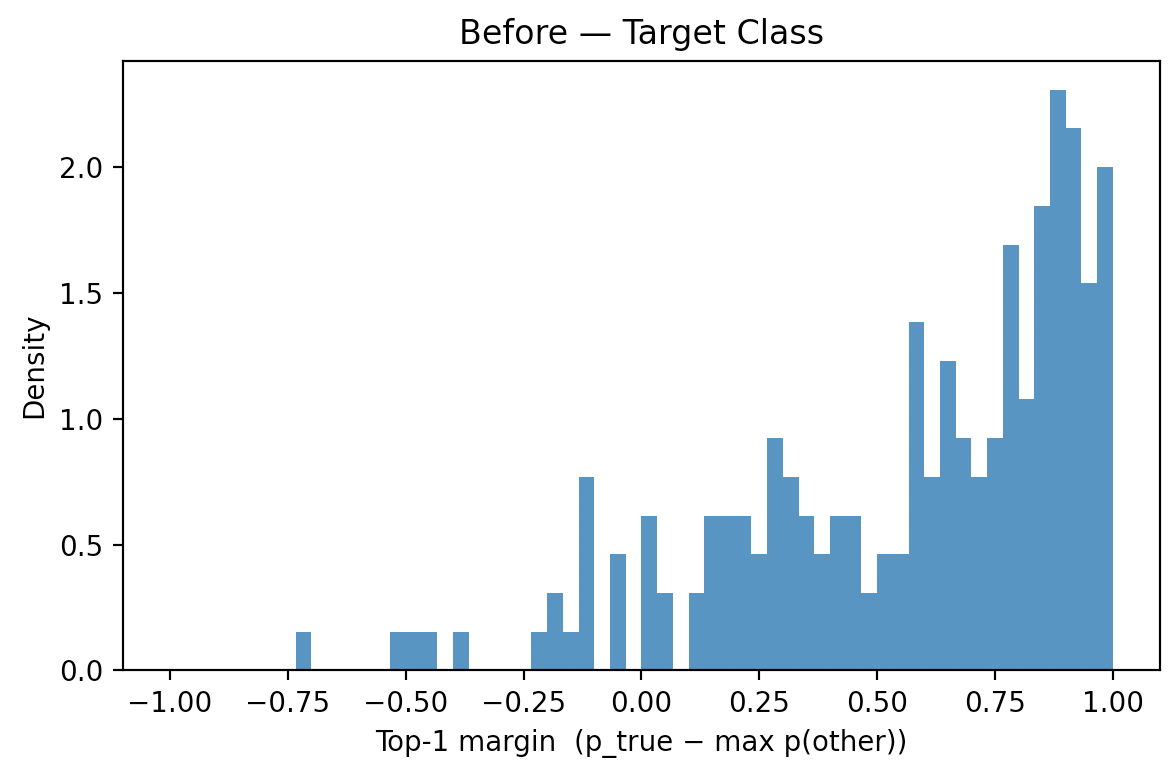}
    \caption{Before Unlearning}
    \label{fig:before_target}
\end{subfigure}
\begin{subfigure}[t]{.5\textwidth}
    \centering
    \includegraphics[width=1.0\linewidth]{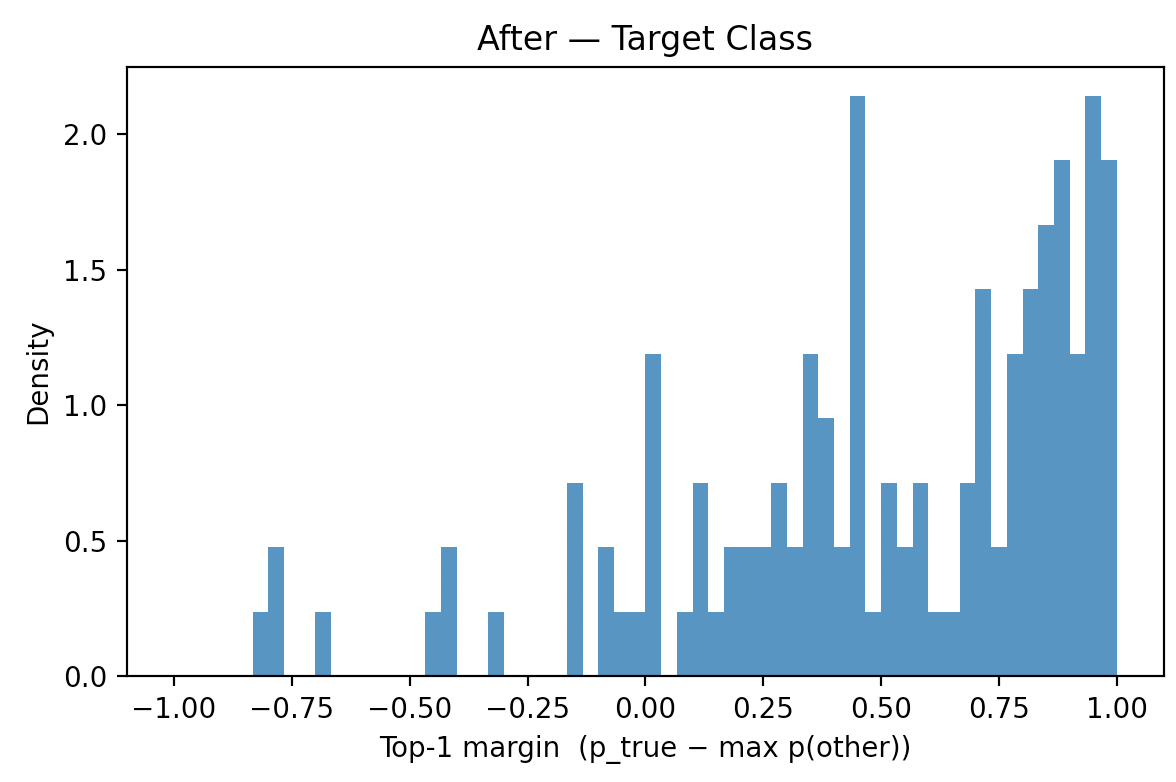}
    \caption{After Unlearning}
    \label{fig:after_target}
\end{subfigure}
\caption{Confidence Distribution on Target Class}
\end{figure}

\noindent \textbf{Target-class documents.} We first examine test examples whose original label is the deleted class $c$. Before unlearning (Fig.~\ref{fig:before_target}), these examples have margins $m(x;c)$ that are mostly large and positive, indicating that the model strongly prefers class $c$ on inputs it has been trained to recognize. After unlearning (Fig.~\ref{fig:after_target}), we treat the Golden Standard reassignment $\tilde y(x)$ (obtained from retraining on $\mathcal{D}_{-c}$) as the reference label and plot $m(x;\tilde y(x))$. The post-unlearning distribution again shows large positive margins: \textbf{the model now assigns high confidence to the next-top-1 labels in $\mathcal{Y}_{-c}$ rather than to the removed class}.

\begin{figure}[hbt!]
\begin{subfigure}[t]{.5\textwidth}
    \centering
    \includegraphics[width=1.0\linewidth]{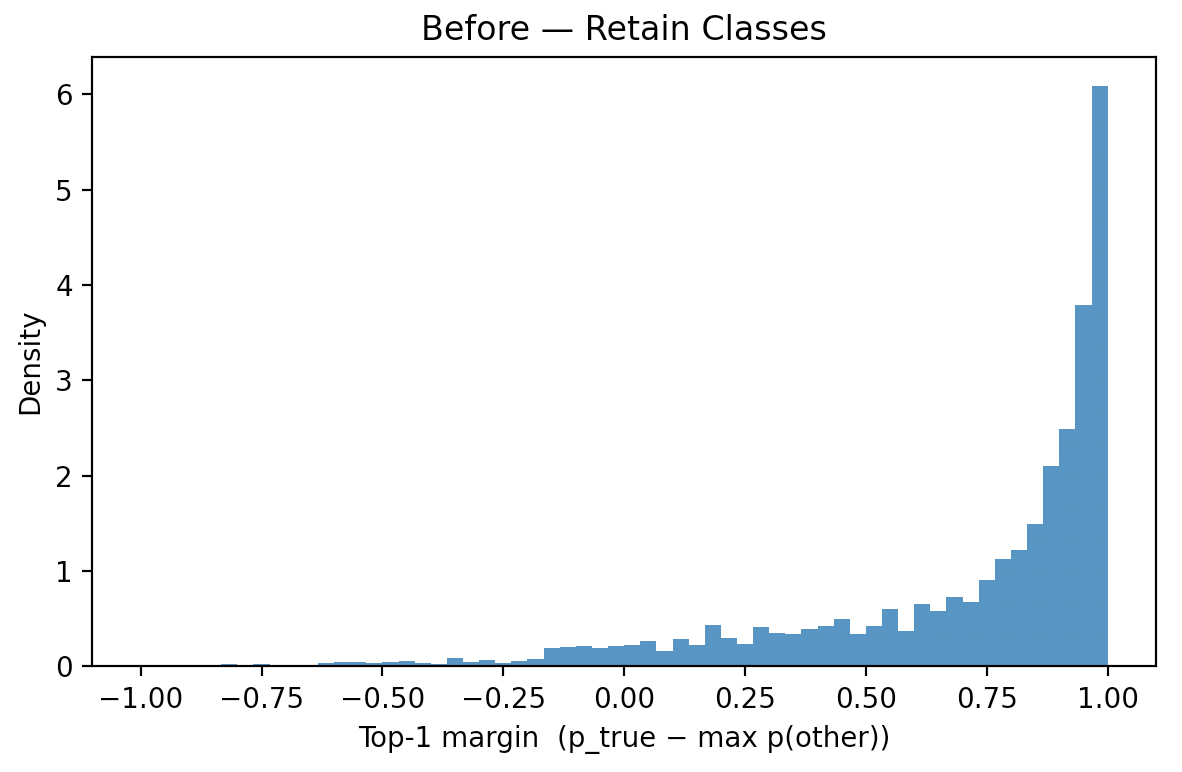}
    \caption{Before Unlearning}
    \label{fig:before_retain}
\end{subfigure}
\begin{subfigure}[t]{.5\textwidth}
    \centering
    \includegraphics[width=1.0\linewidth]{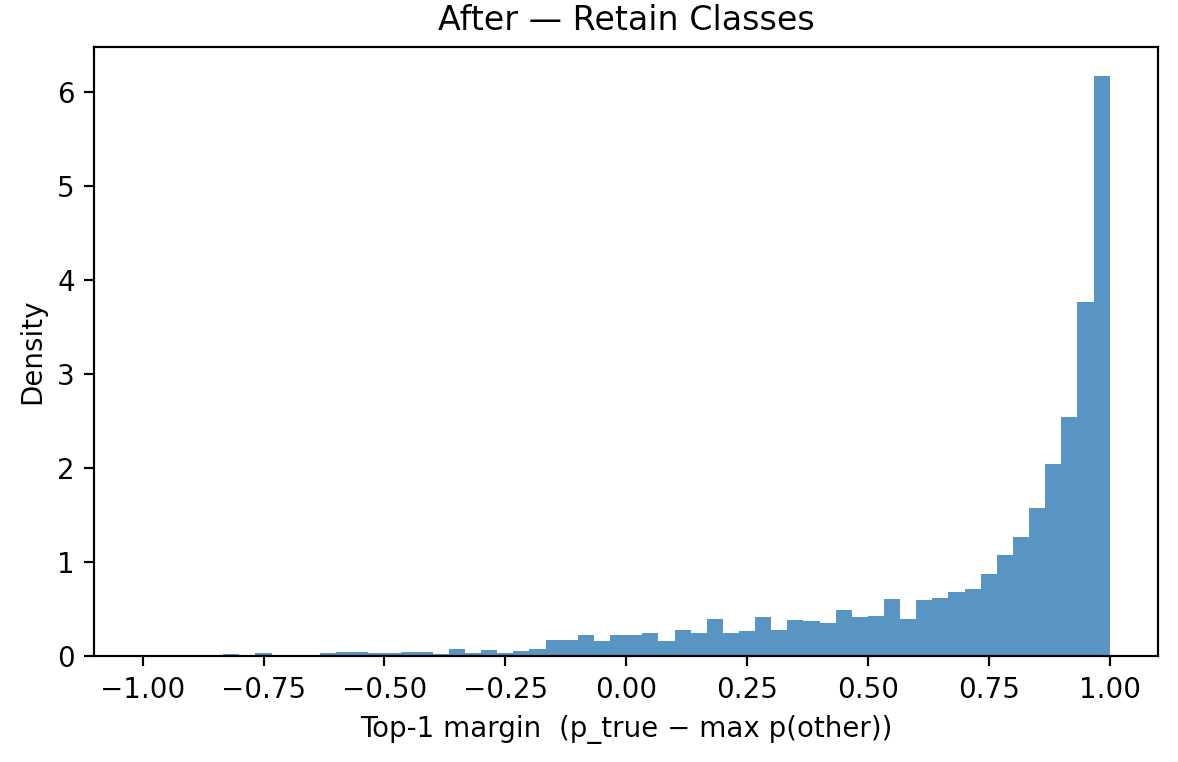}
    \caption{After Unlearning}
    \label{fig:after_retain}
\end{subfigure}
\caption{Confidence Distribution on Retain Classes}
\label{fig:margin_retain}
\end{figure}

\noindent \textbf{Retained-class documents} We next consider test examples with labels in the retained set $\mathcal{Y}_{-c}$. For these, we use the true label as reference and plot $m(x;y)$ before and after unlearning (Fig.~\ref{fig:margin_retain}). In both cases, margins are sharply concentrated near $1.0$. \textbf{The model remains highly confident and correct on retained classes}. Visually, the two curves are almost indistinguishable. Quantitatively, a Kolmogorov--Smirnov test on the before/after distributions yields \textbf{$D = 0.0056$, $p = 1.00$}, indicating no measurable statistical distributional shift after unlearning.  The update therefore targets the removed class without disturbing how confident the model is on the remaining labels.

\subsection{Model Privacy}
\label{sec:privacy}

We evaluate membership-inference (MIA) risk with the fixed shadow-model attacker as described in Section~\ref{sec:metric}. For each method, we report ROC--AUC on (i) documents whose labels are in the retained set ($\text{AUC}_{\mathrm{ret}}$) and (ii) documents whose true label equals the removed class ($\text{AUC}_{c}$). Table~\ref{tab:privacy} summarizes results across datasets.

\begin{table}[ht]
\centering
\caption{Membership-inference ROC--AUC.}
\label{tab:privacy}
\setlength{\tabcolsep}{10pt}
\begin{tabular}{l l c c c}
\toprule
\textbf{Method} & \textbf{Group} & \textbf{20Newsgroups} & \textbf{AG News} & \textbf{DBPedia} \\
\midrule

\multirow{2}{*}{\makecell[l]{Pre-\\Unlearning}}
  & Retain & 0.6635 & 0.5569 & 0.6832 \\
  & Target & 0.6946 & 0.5510 & 0.6794 \\
\midrule

\multirow{2}{*}{\makecell[l]{Random\\Relabeling}}
  & Retain & 0.6346 & 0.5163 & 0.6147 \\
  & Target & 0.5530 & 0.5181 & 0.5228 \\
\midrule

\multirow{2}{*}{\makecell[l]{Hessian\\Reassignment}}
  & Retain & 0.6219 & 0.5235 & 0.6234 \\
  & Target & 0.5109 & 0.5161 & 0.5094 \\
\bottomrule
\end{tabular}
\end{table}

\noindent In this setting, an ROC--AUC of $1.0$ corresponds to a \emph{perfect} attacker that can always distinguish training examples from non-members, and an ROC--AUC of $0.5$ corresponds to a \emph{random-guess} attacker that does no better than flipping a fair coin. Values closer to $0.5$ therefore indicate stronger privacy, while values well above $0.5$ indicate leakage.

Across all three datasets, \textbf{Hessian Reassignment} consistently drives the \emph{Target AUC} (deleted class) toward the random-guess regime of 0.5. $\text{AUC}_c$ drops from roughly $0.69$ (pre-unlearning) and $0.65$ (Random Relabeling) on 20 Newsgroups to $0.51$ with our method, and shows similar behavior on AG News and DBPedia--14. At the same time, the \emph{Retain AUC} (non-deleted classes) remains in the same range as the baselines. This pattern means that the Hessian downweight step is particularly effective at erasing membership
information tied to the deleted class, without materially harming the privacy
of the retained classes.

\subsection{Noise Trade-Offs}
We study the privacy-utility trade-off by injecting Gaussian logit noise at release after our Hessian downweight. For each regularization $C$ and privacy target $\tau$ (desired 
$\text{AUC}_{ret}$), we search the smallest $\sigma^*$ such that 
$\text{AUC}_{ret}\!\le\!\tau$, 
and we report (i) retained-class accuracy at $\sigma^*$ and
(ii) $\log_{10}\sigma^*$. 

\begin{figure}[ht!]
  \centering
  \includegraphics[scale=0.5]{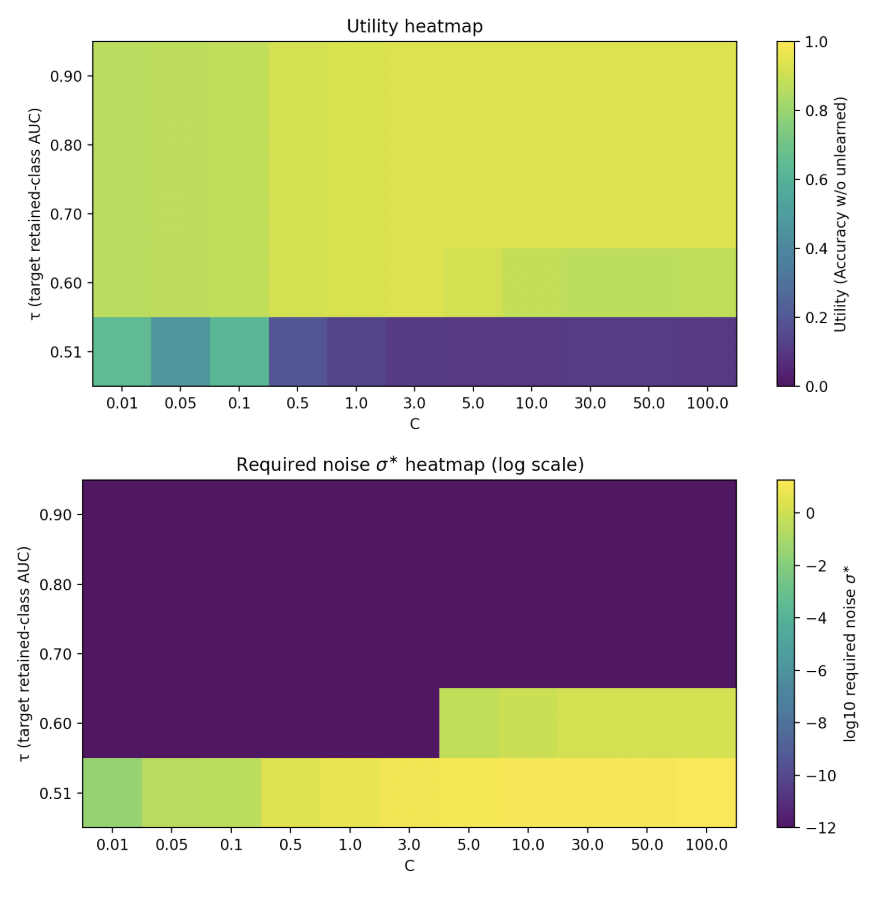}
  \caption{Privacy-utility heatmaps. 
  \emph{Top:} retained-class accuracy attained by the minimum noise $\sigma^*$ that achieves the retain-class $\text{AUC}_{ret}\!\le\!\tau$. \emph{Bottom:} $\log_{10}\sigma^*$ (purple $\approx 0$ noise).}
  \label{fig:noise-heatmaps}
\end{figure}

\noindent Figure~\ref{fig:noise-heatmaps} illustrates our results. For moderate $\tau$ (e.g., $0.6$), $\sigma\approx\!0$ across most $C$ and utility remains maximal. Pushing $\tau$ toward $0.5$ requires nonzero noise chiefly at larger $C$, with a measured but localized utility drop. In all other settings, utility is essentially unchanged, indicating that our method already yields strong privacy on the deleted class without sacrificing performance on retained classes.

\subsection{Model Efficiency}
We measured end-to-end time to produce the released model after a deletion request for both full retraining (\textbf{Gold Standard}) and \textbf{Hessian Reassignment}. 

\begin{figure}[ht!]
  \centering
  \includegraphics[width=0.45\linewidth]{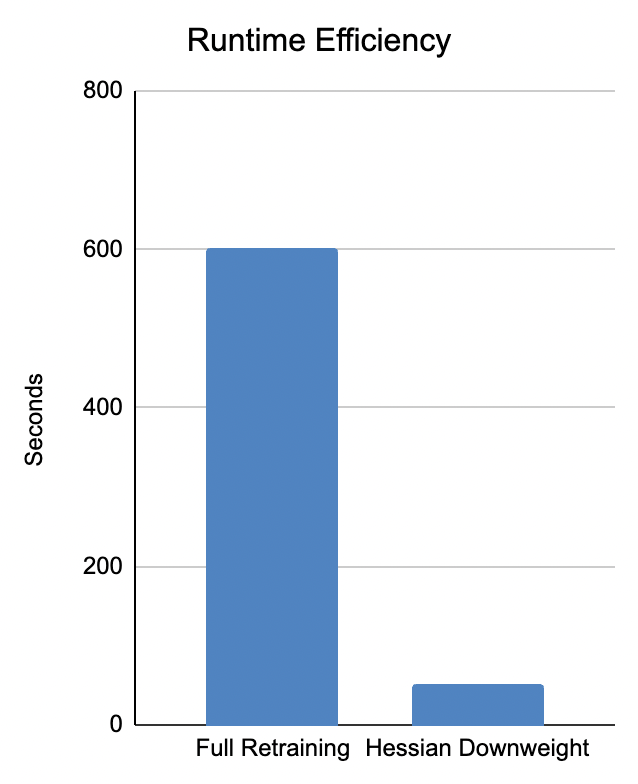}
  \caption{Runtime efficiency}
  \label{fig:efficiency}
\end{figure}

\noindent Figure~\ref{fig:efficiency} compares the  time to produce the released model on DPBedia-14. \emph{Complete Retraining} takes roughly $\sim\!602$ seconds (around 10 minutes), whereas \textbf{Hessian Reassignment} finishes in about $\sim\!53$ seconds. The gain comes from replacing full optimization over the corpus with a single conjugate–gradient solve using Hessian–vector products. Importantly, this acceleration does not sacrifice quality. Sections~\ref{sec:utility} and~\ref{sec:privacy} show retained–class accuracy and target class privacy remaining high. 

\section{Discussion}
This work shows that deterministic next–top-1 reassignment through a single \emph{Hessian downweight} step provides a practical route to class-level unlearning for document classifiers. Empirically, \textbf{Hessian Reassignment} matches \textbf{Golden Standard} (complete retraining) on retained-class utility and consistently surpasses \textbf{Random Relabeling}. The retained-class confidence distribution remains effectively unchanged (e.g., KS $p{=}1.0$ on 20~Newsgroups), while the model is confident in its next top-1 prediction for the target class. Privacy follows suit: membership-inference ROC–AUC on the target class moves toward chance ($\text{AUC}_{c}\!\approx\!0.5$) without sacrificing retained-class accuracy.

The top–1 reassignment preserves the model’s \emph{relative} score ordering among non-target labels, avoiding contradictory supervision that random relabeling introduces. The Hessian downweight then acts as an inexact Newton correction, canceling the deleted class’s first-order influence using the local curvature, yielding a parameter move that respects the existing geometry of the decision boundaries. As a result, no iterative retraining is required.

For multi-class text settings with moderate regularization, the method is a strong default: it achieves near–retraining utility on retained classes and drives the target-class MIA toward $0.5$ without extra noise. When $\text{AUC}_{ret}$ remains above a policy threshold, our heatmaps indicate that only small logit noise is needed to close the gap. 

\section{Future Work}
Looking ahead, several directions appear promising. First, the Hessian downweight step could be adapted to richer text pipelines, including neural networks and transformer models.  Additionally, developing lightweight certificates that quantify how much influence remains after the downweight update would move the approach closer to provable guarantees. Together, these directions would help translate the simplicity of our framework to a wider range of modern NLP systems.

\section{Conclusion}

We presented a simple, efficient algorithm for class-level unlearning in document classification: deterministically reassign deleted-class examples to their next top–1 label and apply a single \emph{Hessian downweight} to remove the deleted class’s first-order influence. Across standard text benchmarks, \textbf{Hessian Reassignment} matches the \textbf{Golden Standard} (complete retraining) on retained-class accuracy, substantially outperforms \textbf{Random Relabeling}, and drives target-class membership–inference AUC toward chance, all while leaving retained-class confidence distributions effectively unchanged.

Conceptually, next–top-1 reassignment preserves the model’s learned ordering among the remaining labels, and the downweight step uses local curvature to make the smallest corrective move consistent with that structure. Practically, this yields a training-free alternative to complete retraining with strong utility–privacy balance. 

\section{Acknowledgement}

I would like to thank the MIT PRIMES program for their support and the opportunity to conduct this research. I am also grateful to my mentor, Mayuri Sridhar, for her extensive guidance and help.

\end{document}